\documentclass[10pt,twocolumn,letterpaper]{article}

\usepackage{cvpr}              
\definecolor{cvprblue}{rgb}{0.21,0.49,0.74}
\usepackage[pagebackref,breaklinks,colorlinks,allcolors=cvprblue]
{hyperref}
\usepackage{multirow}
\usepackage{pifont} 

\usepackage{algorithm}
\usepackage{algpseudocode}
\usepackage[table]{xcolor}
\definecolor{mypink}{rgb}{.99,.91,.95}

\usepackage{ulem}

\definecolor{mycyan}{cmyk}{.1,0,0,0}
\def\confName{CVPR}
\def\confYear{2026}

\title{Learning Surgical Robotic Manipulation with 3D Spatial Priors}
\author{
Yu Sheng\textsuperscript{1,2}, 
Lidian Wang\textsuperscript{1,2},  
Xiaomeng Chu\textsuperscript{3},
Min Cheng\textsuperscript{2}, \\
Bei Hua\textsuperscript{1}, 
Yanyong Zhang\textsuperscript{1}, 
Jiajun Deng\textsuperscript{1}$^\dag$,
Jianmin Ji\textsuperscript{1}$^\dag$\\
$^{1}$University of Science and Technology of China, $^{2}$Tuodao Medical, $^{3}$Yale University\\
{\tt\small \{shengyu724,lidianw,cxmeng\}@mail.ustc.edu.cn}\quad
{\tt\small \{bhua,yanyongz,dengjj,jianmin\}@ustc.edu.cn} \\
{\small $\dag$: Corresponding Author}
}

\begin{document}
\twocolumn[{%
\renewcommand\twocolumn[1][]{#1}%
\maketitle
\begin{center}
    \centering
    \includegraphics[width=.95\textwidth]{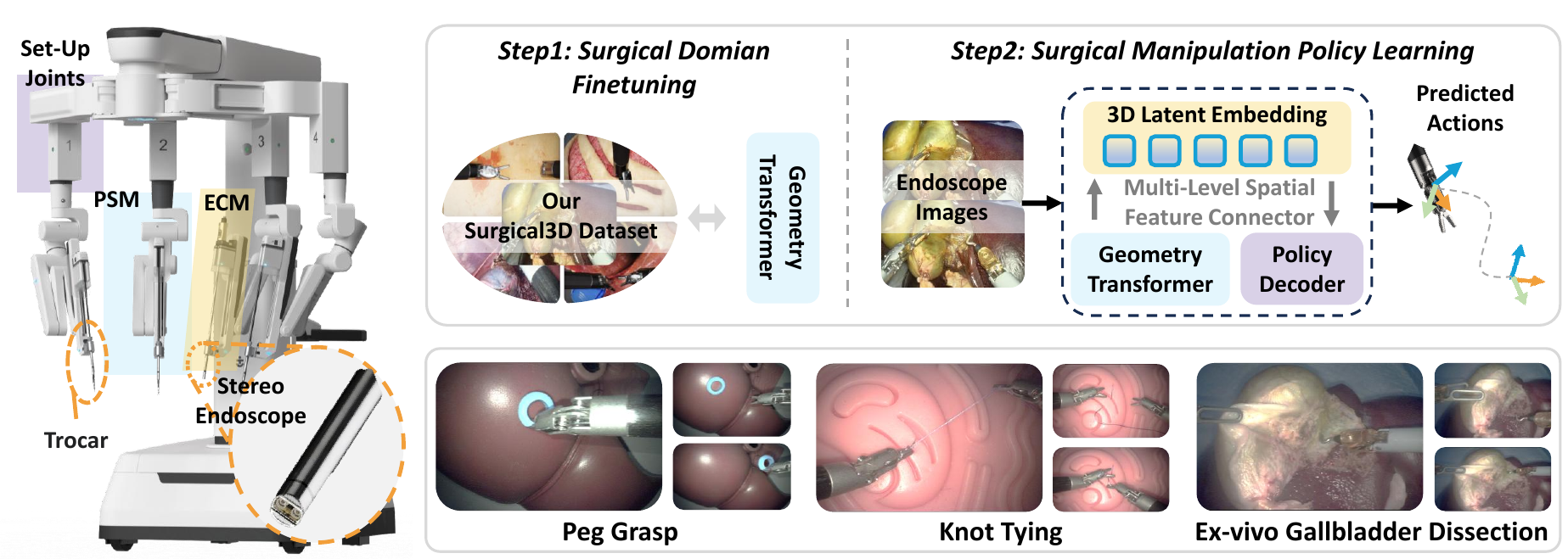}
        \captionof{figure}{We present Spatial Surgical Transformer (SST), a visuomotor policy that empowers surgical robots with spatial intelligence through learned 3D spatial priors. The policy leverages a geometry transformer finetuned on the proposed Surgical3D Dataset to extract robust 3D latent embeddings from stereo endoscopic inputs, coupled with a multi-level spatial feature connector that integrates multi-level 3D latent embeddings capturing both fine-grained details and global context into the policy decoder. We implement SST on a real surgical robot equipped with a stereo endoscopic camera manipulator (ECM) and two patient-side manipulators (PSMs), and evaluate it across three distinct real-world surgical scenes, where it achieves state-of-the-art performance.}
    \label{fig:first-fig}
\end{center}%
}]

\begin{abstract}
Achieving 3D spatial awareness is crucial for surgical robotic manipulation, where precise and delicate operations are required.  Existing methods either explicitly reconstruct the surgical scene prior to manipulation, or enhance multi-view features by adding wrist-mounted cameras to supplement the default stereo endoscopes.  However, both paradigms suffer from notable limitations: the former easily leads to error accumulation and prevents end-to-end optimization due to its multi-stage nature, while the latter is rarely adopted in clinical practice since wrist-mounted cameras can interfere with the motion of surgical robot arms.  In this work, we introduce the Spatial Surgical Transformer (SST), an end-to-end visuomotor policy that empowers surgical robots with 3D spatial awareness by directly exploring 3D spatial cues embedded in endoscopic images. First, we build Surgical3D, a large-scale photorealistic dataset containing 30K stereo endoscopic image pairs with accurate 3D geometry, addressing the scarcity of 3D data in surgical scenes. Based on Surgical3D, we finetune a powerful geometric transformer to extract robust 3D latent representations from stereo endoscopes images. These representations are then seamlessly aligned with the robot's action space via a lightweight multi-level spatial feature connector (MSFC), all within an endoscope-centric coordinate frame.  Extensive real-robot experiments demonstrate that SST achieves state-of-the-art performance and strong spatial generalization on complex surgical tasks such as knot tying and ex-vivo organ dissection, representing a significant step toward practical clinical deployment. The dataset and code will be released.
\end{abstract}    
\section{Introduction}
\label{sec:intro}
Autonomous surgery with surgical robots such as the da Vinci Surgical System represents a highly valuable frontier in clinical practice~\cite{survey}. 
Nevertheless, with the recent progress of visual imitation learning in general-purpose robotic manipulation~\cite{pi0,pi05,open-vla,act,DP,SpatialVLA,GraspCoT}, extending these techniques to surgical scenarios offers a promising direction. However, a key challenge lies in enabling visuomotor policies with 3D spatial awareness, as surgical robots must handle extremely small and delicate structures, such as needles and tissues, with millimeter-level precision.

To tackle this challenge, pioneering works~\cite{suturing_1,suturing_3,suturing_4} typically first reconstruct the 3D surgical scene from stereo endoscopic camera manipulator (ECM) images through optimization-based techniques~\cite{igev,sfm}, and then 
learn and execute surgical manipulation skills on top of the reconstructed scenes~\cite{suturing_2,vppv}.
Such pipelines inevitably accumulate errors during the explicit reconstruction process and cannot be optimized end-to-end due to their multi-stage design.
More recent approaches~\cite{SRT,SRT-H} incorporate wrist-mounted cameras on the patient-side manipulators (PSMs) as a supplement to endoscopic cameras and train an end-to-end visuomotor policy. 
While the additional view offer richer visual information, these methods suffer from two major limitations. 
First, without 3D supervision or geometric priors, the multi-view features fail to capture meaningful geometric cues.
More importantly, wrist cameras are impractical in real surgical procedures~\cite{SRT}, as the trocar (shown in the left of Fig.~\ref{fig:first-fig}) imposes strict spatial constraints on the PSM’s insertion path, preventing instruments equipped with additional cameras from passing through.






Recent advances in feed-forward geometric models have enabled rapid 3D reconstruction~\cite{leroy2024mast3r,wang2024dust3r,wang2025vggt,pixel-splat,transsplat,mvsplat}, producing the latent embedding that captures rich and consistent geometric information. Leveraging these embeddings offers great potential to provide 3D priors for surgical manipulation, alleviating the inefficiency of explicit reconstruction and the hardware constraints of additional sensors. Despite this promise, integrating feed-forward geometric models into autonomous surgery remains highly non-trivial. On the one hand, there is no large-scale public surgical dataset with 3D annotations to-date, and existing 3D reconstruction models are rarely trained on surgical scenes, leading to significant domain discrepancies when applied directly. On the other hand, naively integrating high-capacity pre-trained encoders into visuomotor policy often degrades performance, as their representations are misaligned with task-specific objectives~\cite{act,DP}.

By leveraging feedforward 3D geometric models and addressing the limitations above, we propose a framework that learns 3D spatial priors and integrates them into surgical visuomotor policies.
Our approach is built upon two key components:
(a) We construct Surgical3D, a synthetic dataset with 30K photorealistic stereo endoscopic image pairs and corresponding 3D point maps, filling the gap of publicly available 3D reconstruction data for surgical scenarios.
(b) We introduce \textbf{S}patial \textbf{S}urgical \textbf{T}ransformer (\textbf{SST}), an end-to-end visuomotor policy that leverages 3D spatial priors for precise and generalizable surgical manipulation in clinical settings.
As shown in Fig.~\ref{fig:first-fig} (top right), we first finetune a geometry transformer on the Surgical3D dataset to extract robust 3D latent embeddings from stereo endoscope images, laying a reliable spatial foundation for downstream visuomotor policy learning in diverse surgical environments. We then design a lightweight Multi-Level Spatial Feature Connector (MSFC) that efficiently integrates spatial features capturing both fine-grained details and global context into the policy decoder. Furthermore, since the learned 3D latent embeddings are represented in the endoscope coordinate system, we additionally introduce an endoscope-centric action space to ensure the perception and action are consistently aligned during policy learning and execution. We train SST within an imitation learning framework and deploy it on a real surgical robot, conducting extensive evaluations across tasks ranging from silicone model manipulation to ex-vivo organ dissection, as illustrated in Fig.~\ref{fig:first-fig} (bottom right). SST achieves state-of-the-art performance and strong spatial generalization, showing that learned spatial priors are critical for autonomous surgical systems poised for clinical deployment.


In summary, our main contributions are as follows:
\begin{itemize}
\item We construct Surgical3D, a large-scale synthetic dataset comprising 30K photorealistic stereo endoscope image pairs with precise 3D geometric ground truth, addressing the critical shortage of 3D data in surgical domains.
\item We propose Spatial Surgical Transformer (SST), a visuomotor policy empowered by learned 3D spatial priors. SST capitalizes on a geometry transformer finetuned on Surgical3D to produce high-quality 3D latent embeddings, and a lightweight connector (MSFC) to align multi-level spatial features with the robot’s action space.
\item We deploy SST on a real surgical robot and evaluate it across diverse surgical scenes, achieving state-of-the-art success rates and strong spatial generalization.
\end{itemize}

\section{Related Work}
\label{sec:related}

\begin{figure*}[t]
\centering
 \begin{minipage}{1\linewidth}
    \centering
        \includegraphics[width=1\linewidth]{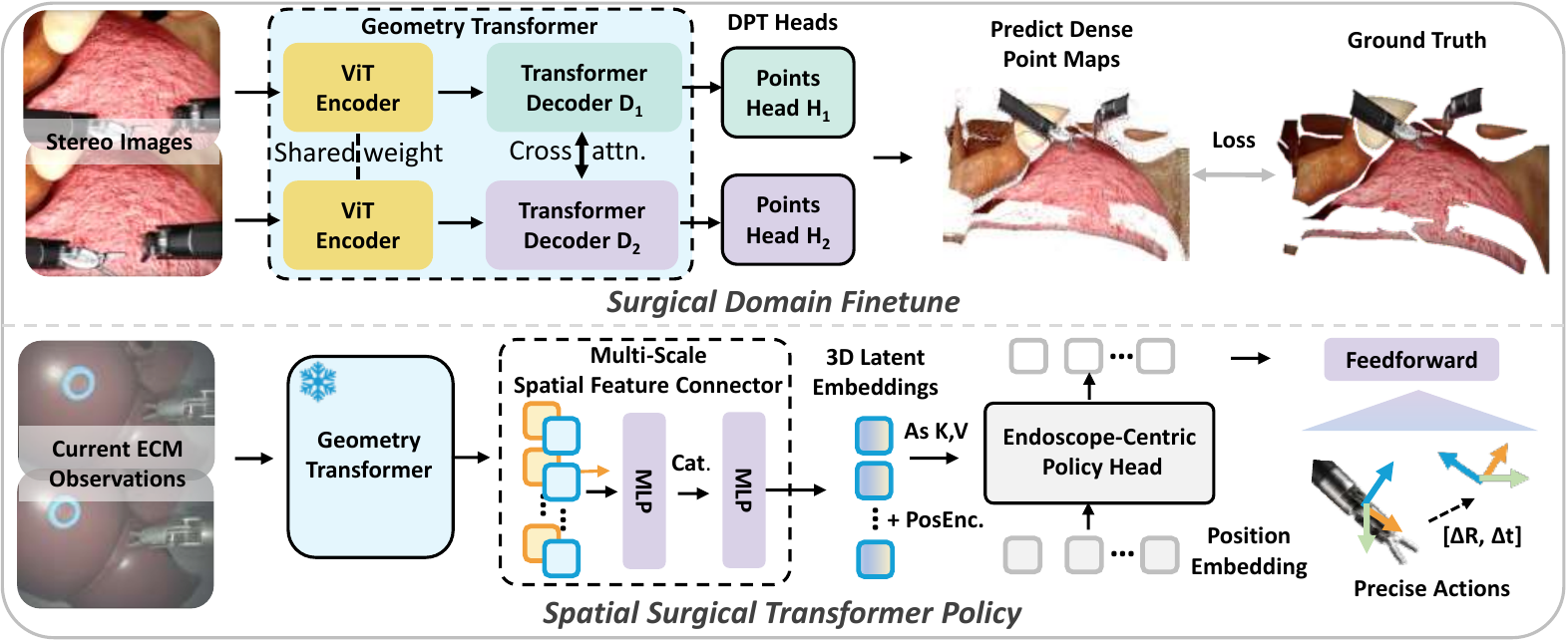}
    \end{minipage}
\caption{\textbf{Pipeline of Our Method.} \textbf{Top:} The geometry transformer is first finetuned on the proposed Surgical3D dataset using a 3D reconstruction objective, enabling the extraction of robust 3D latent embeddings from endoscopic images. \textbf{Bottom:} The geometry transformer is then frozen, while the remaining components are trained to learn surgical manipulation policies with spatial priors from collected demonstrations. A multi-level spatial feature connector (MSFC) is trained to aggregate 3D latent embeddings from multiple geometry transformer blocks and aligns them with the robot’s action space. An endoscope-centric policy decoder generates relative robot actions in the endoscope frame, guided by the learned 3D spatial priors.}
\label{fig:pipeline}
\vspace{-1em}
\end{figure*}

\subsection{Imitation Learning for Robotic Manipulation}
Visual imitation learning plays a central role in robotic manipulation. Early approaches~\cite{bc_1,bc_2} largely relied on behavioral cloning (BC) for task-specific action prediction from observations. Recent advances employ generative modeling techniques such as energy-based models~\cite{energy-based}, diffusion models~\cite{DP}, and variational autoencoders~\cite{act,vae-based}, achieving stronger performance in fine-grained manipulation.
However, these methods often suffer from limited generalization. To address this challenge, large-scale pretrained visuomotor policies such as RT~\cite{RT1}, PI~\cite{pi0,pi05}, and Open-VLA~\cite{open-vla} exploit extensive datasets to enable multi-task capabilities. Meanwhile, several works~\cite{SpatialVLA,vggt-dp,3d-policy1} incorporate 3D information into policy learning to enhance robustness and spatial reasoning. Yet, existing efforts focus predominantly on tabletop environments, while surgical manipulation demands substantially higher precision under limited data and constrained sensing conditions.

\subsection{3D Reconstruction}
Traditional methods such as SfM~\cite{sfm} perform 3D reconstruction through a multi-stage pipeline that involves detecting, matching, and triangulating feature points for an initial sparse reconstruction, followed by dense refinement. The advent of NeRF~\cite{nerf} and 3DGS~\cite{3dgs} has revolutionized this paradigm by introducing differentiable 3D representations, enabling high-quality dense reconstruction and novel-view synthesis. More recent feed-forward approaches~\cite{mvsplat,pixel-splat,pixel-nerf,IBRnet,MuRF,GeoNeRF,transsplat,spatialsplat} learn geometric priors directly from data, achieving rapid reconstruction and consistent 3D feature representations. Notably, DUSt3R~\cite{wang2024dust3r} and MASt3R~\cite{leroy2024mast3r} demonstrate robust reconstruction from unposed image pairs, while VGGT~\cite{wang2025vggt} generalizes this paradigm to more flexible input settings, including single-view, multi-view, and video data. Leveraging internet-scale training, these models offer fast inference, strong generalization, and consistent 3D representations. In contrast, surgical-scene reconstruction remains constrained by the scarcity of publicly available 3D-annotated datasets. Existing methods are thus limited to geometry-based approaches~\cite{trad_3d_1,trad_3d_2,trad_3d_3,trad_3d_4} or refinement-based networks~\cite{dl_3d_1,dl_3d_2,dl_3d_3,dl_3d_4}, both of which rely on expensive per-scene optimization and are impractical for real-time robotic policy execution. This lack of in-domain 3D data also hampers the transferability of feed-forward foundation models such as MASt3R and VGGT to surgical imagery, resulting in suboptimal performance when applied directly.


\subsection{Autonomous Surgery}
Traditional autonomous surgical systems are typically designed for specific tasks and environments, such as suturing~\cite{suturing_1,suturing_2,suturing_3,suturing_4} or endoscope control~\cite{endo_ctrl_1,endo_ctrl_2}, and thus lack the ability to generalize across tasks. Subsequently, imitation learning has been applied to simple surgical operations~\cite{imitatial_surgical_1,imitatial_surgical_2,imitatial_surgical_3}, and the SRT series~\cite{SRT,SRT-H} has successfully extended imitation learning to more complex procedures such as knot tying.
Building on this line of work, our approach eliminates the need for multi-stage pipelines or additional sensors such as wrist cameras, resulting in a real-time, end-to-end visual imitation learning policy that achieves comparable or superior performance across diverse surgical tasks.
\begin{figure*}[t]
\centering
\begin{minipage}{1\linewidth}
    \centering
        \includegraphics[width=1\linewidth]{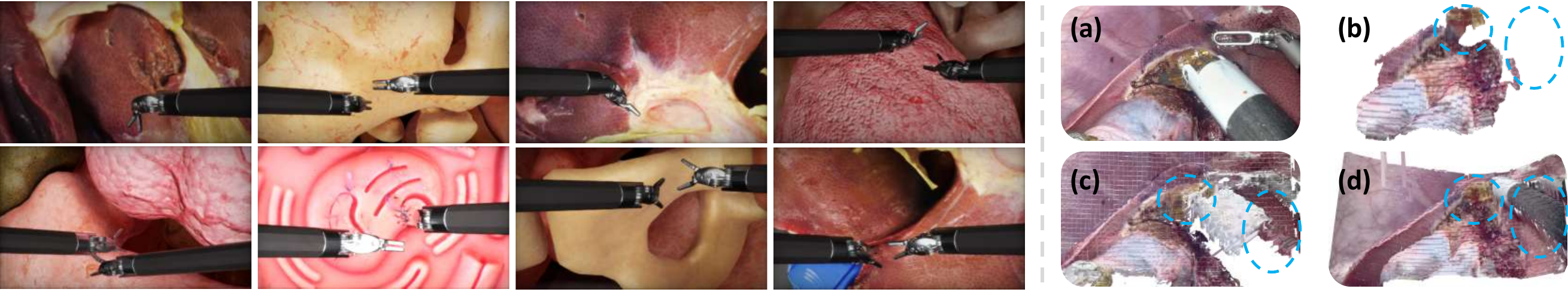}
    \end{minipage}
\caption{\textbf{Left:} Samples from our Surgical3D dataset. We utilize diverse 3D surgical assets to generate highly realistic and varied synthetic surgical scenes. \textbf{Right:} The figure illustrates the reconstruction results from MASt3R under three finetuning configurations.
(a) One example of a stereo endoscopic image captured in a real in-vivo surgical scene, used as input.
(b) The original MASt3R fails to reconstruct both the patient-side manipulator (PSM) and the organ surface.
(c) When finetuned solely on synthetic data, the organ reconstruction remains coarse and the PSM geometry is still incomplete.
(d) When finetuned on a combination of synthetic and real data, the model achieves more accurate reconstruction of PSM instruments and better organ completeness.}
\label{fig:mix_datasets}
\vspace{-1em}
\end{figure*}

\section{Method}
Our goal is to learn a policy $\pi_{\theta}$, parameterized by $\theta$, that maps the current observation $o_t$ and proprioceptive state $x_t$ to the next-step robot action $a_t$ for a given task, using a set of demonstrations $D = { \tau_1, \dots, \tau_n }$ containing $n$ trajectories. Each trajectory $\tau_i = \{ (o_1, x_1, a_1), \dots, (o_T, x_T, a_T) \}$ consists of $T$ time steps. As illustrated in Fig.~\ref{fig:pipeline}, our policy consists of three components: (1) A geometry transformer (Section~\ref{3d_recon_net}) finetuned on the Surgical3D dataset described in Section~\ref{data} with a 3D reconstruction objective, which extracts robust 3D latent embedding from stereo endoscopic observations;
(2) A Multi-Level Spatial Feature Connector (MSFC) that aligns the learned 3D latent embeddings with the action space according to task-specific requirements (Section~\ref{3d_inject}); and (3) An endoscope-centric policy decoder that predicts the robot’s relative actions in the endoscope coordinate system (Section~\ref{3d_Policy}).

\subsection{Surgical3D Dataset}
\label{data}
Existing surgical robotic systems typically rely solely on a stereo endoscopic camera for perception, and the highly confined surgical environment (organ-to-camera distances often fall below $10cm$) exceeds the sensing range of most 3D sensors, making datasets with 3D annotations in surgical scenes extremely scarce. To tackle this limitation, we construct a large-scale synthetic surgical dataset using NVIDIA Omniverse, termed Surgical3D. This dataset mitigates the scarcity of endoscopic surgical data by providing detailed 3D annotations, including depth maps, point clouds, and camera extrinsic parameters. To enhance diversity, we apply domain randomization by varying stereo endoscope baselines, camera intrinsics and extrinsics, lighting conditions, and tissue textures, thereby simulating real-world variability, as illustrated in Fig.~\ref{fig:mix_datasets}. Surgical3D integrates two categories of 3D assets: (1) open-source~\cite{ASSETS} human full-body organ models (8 types) and surgical instrument assets that provide substantial diversity but limited anatomical realism, and (2) 10 realistic 3D meshes captured from real organs using an iPad scanner. These assets collectively generate 30k high-resolution (1920 × 1080) stereo image pairs with corresponding depth maps for training the surgical 3D geometry transformer.

Despite this, a domain gap still remains between synthetic and real surgical scenes due to differences in organ morphology and intra-abdominal lighting. Models trained solely on synthetic data generalize poorly to real settings, as shown in Fig.~\ref{fig:mix_datasets}. To mitigate this gap, we generate pseudo-labels for unlabeled real data (e.g., live surgical recordings). Specifically, we finetune VGGT’s point prediction heads on synthetic data and use the refined model to infer depth maps for real data. To ensure label reliability, only regions with high confidence scores are retained. As shown in Fig.~\ref{fig:mix_datasets}, the hybrid dataset substantially improves the robustness and generalization of our geometry transformer across both synthetic and real domains.

\subsection{Surgical Geometry Transformer}
\label{3d_recon_net}
The geometry transformer aims to map raw stereo endoscopic images into compact 3D latent embeddings. We carefully analyze the unique characteristics of surgical imagery: unlike natural scenes, organ surfaces in surgical environments are often textureless or exhibit highly repetitive patterns, rendering traditional feature-matching methods~\cite{colmap,igev,foundationstereo} unreliable. Furthermore, the extremely narrow baseline between the two cameras in endoscope causes geometry-based methods~\cite{mvsplat,trad_3d_1,trad_3d_2,trad_3d_3,trad_3d_4} to accumulate significant depth errors even from minor pixel misalignments. To address these challenges, we adopt MASt3R~\cite{leroy2024mast3r}, a recent feed-forward 3D reconstruction model that infers dense 3D points from paired images without relying on camera parameters or feature matching, as the prototype for our geometry transformer. This also enables us to leverage MASt3R’s internet-scale pretraining to obtain robust 3D latent embeddings. While the recent popular vision foundation model VGGT~\cite{wang2025vggt} can also capture finer 3D geometric information, its heavy architecture limits real-time deployment on robotic platforms and can induce motion jitter. In contrast, MASt3R, being more lightweight, is sufficiently robust to handle raw stereo images in surgical scenes after finetuned on our Surgical3D dataset. The model is built entirely on ViT structures. Input stereo endoscopic images are first patchified and flattened into token sequences, which are processed by a shared encoder across all views. The resulting tokens are then fed into a ViT-based decoder~\cite{leroy2024mast3r,wang2024dust3r,spatialsplat}, where cross-attention captures spatial relationships and aggregates information across views.

During the geometry transformer finetuning phase, tokens from the decoder are passed through DPT heads~\cite{dpt} to regress dense 3D point maps in endoscope coordinate. Denote the $\hat{X}^{1,1}$ and $\hat{X}^{2,1}$ as the ground-truth point maps for valid pixels $D^1, D^2 \in
\{1 . . . W \} \times \{1 . . . H\}$ in the endoscopic views, the regression loss is defined as Eq.~\ref{reg-loss}.
\begin{equation}
\small
\label{reg-loss}
    L_{reg}(v,i) = \sum_{v=\{ 1,2\}}\sum_{i\in D^v} \|\frac{1}{z}X^{v,1}_i - \frac{1}{\hat{z}}\hat{x}^{v,1}_i \|
\end{equation}
Where $z$ and $\hat{z}$ are scale factors used to resolve the scale ambiguity between predictions and ground truth by normalizing the corresponding point maps. To address regions with minimal texture, which are common in surgical scenes and particularly challenging for 3D structure regression, we adopt a confidence-aware approach following~\cite{leroy2024mast3r}, predicting a per-pixel confidence score $C^{v,1}_i$. The overall training objective is then defined as:
\begin{equation}
\small
\label{conf-loss}
    L_{conf} = \sum_{v=\{ 1,2\}}\sum_{i\in D^v}C^{v,1}_iL_{reg}(v,i) -\alpha \log C^{v,1}
\end{equation}

\subsection{Multi-Level Spatial Feature Connector}
\label{3d_inject}
In this section, we explore how to effectively connect the spatial priors with policy decoder. Directly using explicit 3D point maps as policy inputs can introduce significant errors due to imperfect reconstruction and inherent scale ambiguity. Moreover, naive implicit approaches such as substituting the original image encoder with a more powerful pretrained visual encoder, often provide limited benefits or even degrade performance~\cite{act,DP}. These challenges call for a principled integration strategy capable of effectively aligning geometric priors with robotic actions.

We propose a Multi-Level Spatial Feature Connector (MSFC) that aggregates 3D latent embeddings from both lower and higher transformer layers and aligns them with the action feature space. The multi-level design is motivated by the observation in~\cite{dpt}, which demonstrates that different transformer layers capture features at varying levels of abstraction: lower layers encode fine-grained local details, while higher layers capture more global contextual information. Such multi-level cues are essential for surgical manipulation, where robots must simultaneously reason about precise object locations and overall motion directions. As illustrated in Fig.~\ref{fig:pipeline}, latent embeddings from four decoder layers of our geometry transformer are first projected into a lower-dimensional space for compactness. We then concatenate embeddings from different layers along the feature dimension and align them to the action space using a lightweight MLP. The aligned latent embeddings subsequently perform cross-attention with positional embeddings to generate robot actions. Notably, the four selected layers correspond to those whose embeddings are fed into the DPT heads during in-domain finetuning, as these embeddings are responsible for predicting accurate point maps and thus contain the most informative geometric cues.

\begin{figure*}[t]
\centering
 \begin{minipage}{1\linewidth}
    \centering
        \includegraphics[width=1\linewidth]{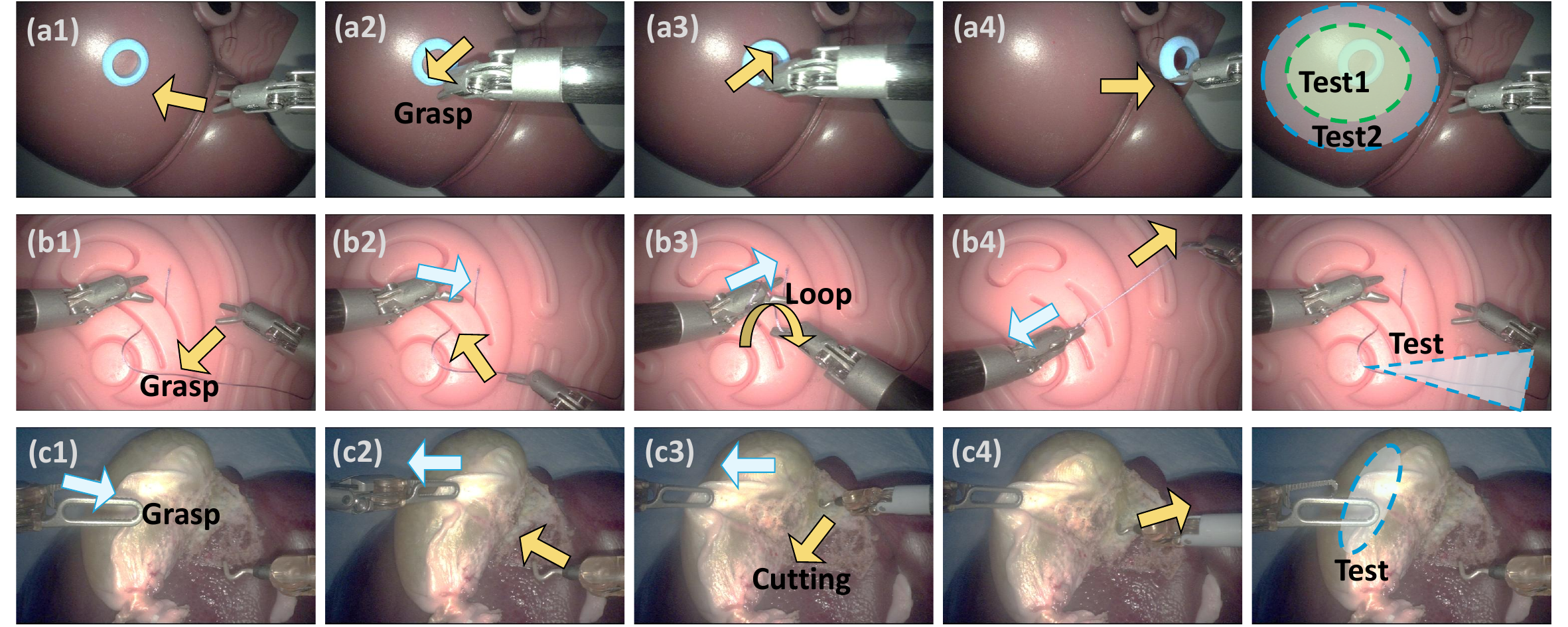}
    \end{minipage}
\vspace{-1em}
\caption{\textbf{Visualization of Experimental Settings.} Yellow and blue arrows denote the approximate motion directions of the right and left arms, respectively. \textbf{Top:} Peg pickup task. A total of 180 trajectories were collected, with roughly 120 in the green region and 60 in the blue region for training. \emph{Test1} and \emph{Test2} correspond to evaluations in these respective areas. \textbf{Middle:} Knot tying task. The suture tail was randomly positioned within the blue region during data collection, and evaluations were performed under the same condition. (b1) and (b3) show the grasping and looping actions. \textbf{Bottom:} Ex-vivo gallbladder dissection task. Grasp points on the gallbladder were sampled from varying positions within the blue region during data collection, and all methods were evaluated under the same setup. (c1) and (c3)  depict the grasping and cutting actions. \textbf{Supplementary videos} provide detailed views of the task execution.}
\label{fig:exp_setting}
\vspace{-1em}
\end{figure*}

\subsection{Endoscope-Centric Policy Decoder}
\label{3d_Policy}
Since the 3D latent embeddings learned in Sec.~\ref{3d_recon_net} are defined in the endoscope coordinate frame, we transform the action space into the same frame, ensuring that the policy operates entirely within a unified endoscope-centric representation. We next describe the action space design and the policy decoder architecture

\vspace{2pt}\noindent\textbf{Action Space.}
A key difference between surgical robots and generalist robots lies in the absence of accurate forward kinematics for surgical systems~\cite{SRT, survey}. As shown in Fig.~\ref{fig:first-fig}, the PSM includes Set-Up Joints (SUJ) that rely solely on potentiometers for joint sensing, making their measurements inherently imprecise. These factors make it infeasible to learn control policies based on absolute joint states or end-effector poses, as commonly done in general-purpose robot learning. To mitigate these effects, we adopt a relative pose representation, similar to~\cite{SRT}. Specifically, record end-effector poses $E^i = (R^i,tr^i) \in SE(3), i\in \{left, right \}$, where $R^i \in SO(3)$ is the rotations matrix and $tr^i \in \mathbb{R}^3$ is the translation vector. And then compute their differences between consecutive frames as:
\begin{equation}
\small
\label{relativa-action}
    \begin{split}
        a_t &= \{E^i_{t+1} \ominus E^i_t\} \\
        &= \{(tr^i_{t+1} - tr^i_t, (R^i_t)^TR^i_{t+1})\}, i\in \{left, right \}
    \end{split}
\end{equation}
Rotation differences are subsequently expressed as Euler angles, which we found to be more conducive to learning than rotation matrix representations in our experimental setting. The gripper is represented by its absolute jaw angle, as it is independent of forward kinematics. This yields a 7-dimensional action space per arm (translation [3] + rotation [3] + jaw angle [1]). All transformations are represented in the endoscope coordinate frame.

\vspace{2pt}\noindent\textbf{Policy Decoder}
Our policy decoder consists of $L$ transformer blocks, each comprising a self-attention layer and a cross-attention layer to effectively integrate the 3D latent embeddings with the action tokens. The policy decoder takes a fixed number of learnable positional embeddings as input. The positional encodings of the 3D latent embeddings are derived by first reshaping the embeddings into image-like tensors, concatenating the embeddings from the two stereo images along the width dimension, and subsequently applying fixed 2D positional encodings~\cite{Image-Transformer}. Since surgical tasks demand smooth and stable motion, directly predicting the next action often leads to jittery trajectories. To address this, we adopt the Action Chunk Transformer (ACT)~\cite{act} framework, which predicts the next $k$ actions based on the current observation. The executed action is then computed as the weighted average of these predicted actions. We employ an exponential weighting scheme $w_i = \exp(-m \cdot i)$, where $i$ denotes the $i$-th future action and $m$ is a hyperparameter controlling the influence of later actions. The policy is trained in an end-to-end manner to minimize the mean squared error (MSE) between the predicted and ground-truth actions:
\begin{equation}
\small
\label{MSE-loss}
    L_{MSE} = MSE(\hat{a}_t,\pi_{\theta}(o_t, x_t))
\end{equation}

\section{Experiments}
\label{sec:exp}

\begin{figure*}[t]
\centering
 \begin{minipage}{1\linewidth}
    \centering
        \includegraphics[width=1\linewidth]{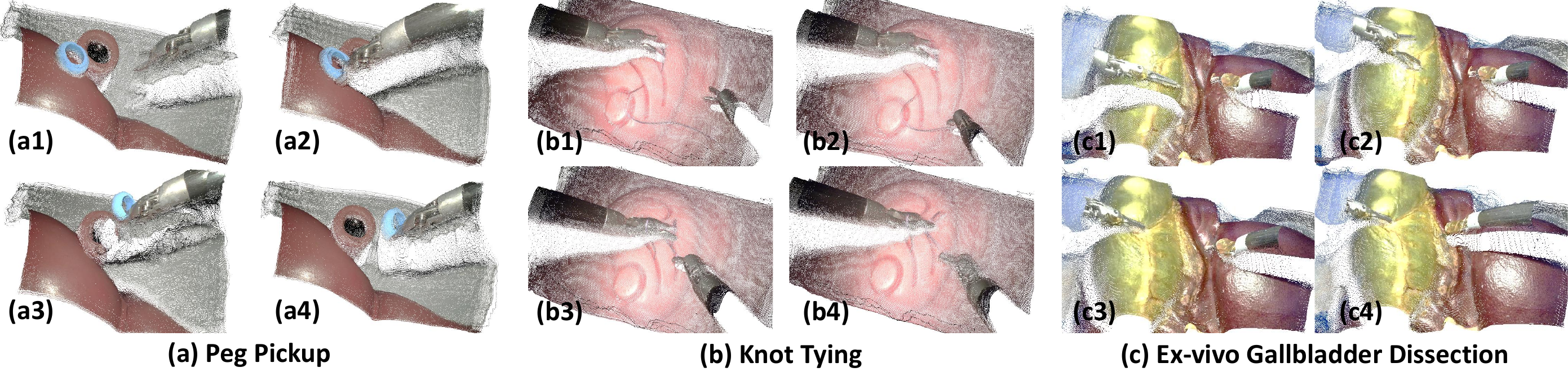}
    \end{minipage}
\vspace{-1em}
\caption{\textbf{Qualitative Results of Intermediate 3D Reconstruction.} Subscripts at the lower left of each image indicate the manipulation step. Tokens from the finetuned geometry transformer are fed into DPT heads to generate intermediate 3D reconstructions. The results demonstrate that the geometry transformer effectively extracts 3D cues from current endoscopic observations across various test tasks without task-specific training. More details are provided in the \textbf{supplementary videos}.}
\label{fig:Qualitative-results}
\vspace{-0.3em}
\end{figure*}

\begin{table*}[t]
\small
    \setlength{\abovecaptionskip}{0.3em}
    \centering
    \resizebox{\linewidth}{!}{
    \begin{tabular}{{l|c|cc|ccc|ccc}}
        \hline
        \hline
         \rowcolor[gray]{.92} & & \multicolumn{2}{c|}{Peg Pickup} & \multicolumn{3}{c|}{Knot Tying} & \multicolumn{3}{c}{Gallbladder Dissection}\\
        \rowcolor[gray]{.92} Methods & Setting  & Test1 & Test2 & Grasp & Loop & Whole Task  & Grasp & Dissection & Whole Task\\
        \hline
        SRT~\cite{SRT} & w/. wrist cams & 10/10 & 6/10 & 10/10 & 3/10 & 2/10 & - & - & - \\
        ACT~\cite{act} & w/o. wrist cams & 9/10 & 2/10 & 4/10 & 0/10 & 0/10 & 0/10 & 0/10 & 0/10 \\
        DP~\cite{DP} & w/o. wrist cams & 10/10 & 1/10 & 5/10 & 1/10 & 1/10 & 0/10 & 0/10 & 0/10 \\
        \rowcolor{mycyan} SST (Ours) & w/o. wrist cams & 10/10 & 8/10 & 10/10 & 7/10 & 7/10 & 10/10 & 6/10 & 6/10 \\
        \hline
        \hline
    \end{tabular}
    }
\caption{\textbf{Success Rates on Three Surgical Tasks Across Various Methods.} 
}
\vspace{-0.3em}
\label{table:main-results}
\vspace{-1em}
\end{table*}

\begin{table}[t]
\small
    \setlength{\abovecaptionskip}{0.3em}
    \centering
    \resizebox{\linewidth}{!}{
    \begin{tabular}{{l|cc|ccc}}
        \hline
        \hline
         \rowcolor[gray]{.92} & \multicolumn{2}{c|}{Peg Pickup} & \multicolumn{3}{c}{Knot Tying}\\
        \rowcolor[gray]{.92} Config  & test1 & test2 & Grasp & Loop & Whole Task\\
        \hline
        w/o ToS & 2/10 & 0/10 & 0/10 & 0/10 & 0/10 \\
        \rowcolor{mycyan} w/ ToS (Ours) & 10/10 & 8/10 & 10/10 & 7/10 & 7/10\\
        \hline
        \hline
    \end{tabular}
    }
\caption{\textbf{Effectiveness of Finetuning Geometry Transformer on Surgical3D Dataset.} ``ToS" indicates that the geometry transformer was trained on the Surgical3D dataset.}
\vspace{-1em}
\label{table:3D-pretraining}
\end{table}

Since no public benchmark exists for surgical robotic manipulation, we deploy SST on the Torin surgical robot (as shown in Fig.~\ref{fig:first-fig}) and conduct a systematic evaluation across three representative real-world tasks: peg pickup, knot tying, and ex-vivo gallbladder dissection, as detailed in Fig.\ref{fig:exp_setting}. This benchmark encompasses single- and dual-arm manipulation, objects with varying sizes and rigidity, both flat and irregular 3D geometries, and includes experiments on silicone organ models as well as real organs. SST consistently achieves comparable, and in some cases superior, performance relative to state-of-the-art methods in both success rate and spatial generalization. In the following sections, we present the implementation details, task setups, and key experimental findings.

\subsection{Experiment Setup} 
\vspace{2pt}\noindent\textbf{Implementation Details.}
We adopt a ViT-Large with a patch size of 16 as the geometry transformer, initialized with MASt3R~\cite{leroy2024mast3r} pre-trained weights. The policy decoder consists of 12 transformer decoder layers with a hidden dimension of 768. For the baseline methods SRT and ACT, we retain their original architectures, four transformer encoder layers and seven decoder layers, each with a 768-dimensional hidden size. Both of us adopt a action chunk of 100 and weighting scheme of $m=0.1$. For Diffusion Policy, we follow the default configuration in~\cite{DP}. Note that for SRT, we incorporate additional wrist cameras, designed to be sufficiently compact to closely replicate the system’s original configuration, whereas all other methods are trained solely on endoscopic images.

\vspace{2pt}\noindent\textbf{Data Collection and Task Setup.}
We collect 180 trajectories for peg pickup, 200 for knot tying, and 200 for ex-vivo gallbladder dissection over multiple days. Peg pickup and knot tying are performed inside a dome simulating the human abdomen to ensure environmental stability, while the ex-vivo gallbladder dissection is conducted in an open environment. We exclude SRT from the gallbladder dissection task, as fluid elements (e.g., water or blood) can damage its wrist cameras. Further details of the experimental setup are provided in the supplementary material.

\vspace{2pt}\noindent\textbf{Evaluation Protocol.}
All methods are trained for 100 epochs, and the final checkpoint is used for evaluation. For the peg pickup task, success is defined as grasping and lifting the peg for at least three seconds. The knot-tying task is divided into two subtasks: grasping, where both arms must successfully grasp the suture, and looping, where the right arm loops the suture once around the left arm. In the ex-vivo gallbladder dissection task, success requires first grasping and pulling the gallbladder leftward, followed by positioning the right arm at the gallbladder–liver interface to perform the cutting motion. For repeatability, we only execute the cutting motion without dissecting the tissue. Each task is evaluated over 10 independent trials.

\subsection{Main Results}
Our experiments are designed to evaluate:
(1) The performance of SST across different tasks in comparison with existing approaches;
(2) The degree to which our method enhances spatial generalization over prior methods; and
(3) The ability of the geometry transformer to extract robust 3D latent embeddings across diverse and unseen scenes.

\vspace{2pt}\noindent\textbf{Success Rates Across Multi-Tasks.} As shown in Tab.~\ref{table:main-results}, across all tasks, SST consistently achieves comparable or superior performance to state-of-the-art methods, while operating under a more practical setting that relies solely on endoscopic images. In contrast, directly training ACT or Diffusion Policy using only endoscopic inputs leads to a significant performance drop, performing reasonably well only on the simplest peg pickup task and failing entirely on more complex tasks such as ex-vivo dissection. Notably, SST demonstrates superior performance in ex-vivo experiments, despite the evaluation gallbladder differing substantially from those in the Surgical3D dataset. The relatively lower success rate in the dissection subtask is due to occasional difficulty in precisely locating the edge between the gallbladder and liver, a challenge even for human operators. SRT performs well on peg pickup tasks in both settings, highlighting the benefit of wrist cameras. It is worth noting that, even without wrist cameras, SST achieves comparable grasp performance and superior loop manipulation performance to SRT in the knot-tying task. This further demonstrates that leveraging stable 3D geometry is significantly more effective than training a policy from scratch, highlighting the crucial role of robust 3D latent embeddings in achieving reliable and precise surgical manipulation.

\vspace{2pt}\noindent\textbf{Spatial Generalization.}
In the peg pickup task, we deliberately employ a silicone liver model with irregular 3D geometry. Consequently, the \underline{Test2} region introduces both an expanded pickup range and substantial depth variation due to the non-planar surface. While all methods achieve high success rates within the constrained green test region, performance diverges markedly in the broader blue region. Methods such as ACT tend to attempt pickup at the green circle even when the peg is placed in the blue area, whereas SST accurately adapts to the peg’s actual position, successfully grasping even at the region’s edge. SRT also exhibits a certain degree of spatial generalization; however, it fails when the peg is placed at the upper or left areas in the \underline{Test2} setup. Interestingly, we observed that when grasping pegs located at left edge, the wrist camera almost completely occludes the endoscopic view, which likely contributes to the relatively low success rate. These results highlight SST’s superior capability to perceive and reason over complex 3D structures, enabled by its powerful geometry transformer finetuned on Surgical3D.
\begin{figure}[t]
\centering
\begin{minipage}{1\linewidth}
    \centering
        \includegraphics[width=0.95\linewidth]{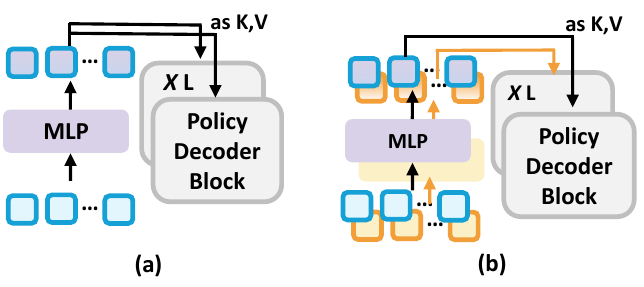}
    \end{minipage}
\vspace{-1em}
\caption{\textbf{Alternative Designs of Spatial Connectors.} (a) Last-Layer Feature Connector. (b) Multi-Layer Separate Connector
}
\label{fig:injector}
\vspace{-1em}
\end{figure}

\vspace{2pt}\noindent\textbf{Robustness of 3D Latent Embeddings.}
We selected three tasks spanning distinct visual domains, including both real organs and silicone organ models, all of which were unseen during geometry transformer finetuning stage, to evaluate its cross-scene generalization. Despite substantial domain shifts, SST consistently maintains stable manipulation performance across all settings, demonstrating that its geometry transformer can extract robust 3D latent embeddings without task-specific finetuning. Qualitative visualizations of intermediate 3D reconstruction results (not utilized by the policy) in Fig. \ref{fig:Qualitative-results} further highlight the geometry transformer’s strong generalization capability across diverse surgical scenes.

\subsection{Ablation Study} 
Our ablation study examines three key design choices: (1) the effect of finetuning the geometry transformer on the Surgical3D dataset, (2) the efficiency of MASt3R as geometry transformer,  and (3) the effectiveness of the Multi-Level Spatial Feature Connector (MSFC). Experiments are conducted on the peg pickup and knot tying tasks, with each task evaluated over 10 trials.

\vspace{2pt}\noindent\textbf{Effective of Finetuning on Surgical3D.}
As shown in Tab.~\ref{table:3D-pretraining}, on the peg pickup task, SST without finetuning geometry transformer on Surgical3D performs significantly worse. The policy can roughly locate the peg position but often grasps with noticeable spatial offsets, likely due to the it’s inability to extract reliable 3D cues from surgical scenes. In the knot tying task, it fails to learn meaningful behaviors from demonstrations, producing aimless actions. In contrast, finetuning on our Surgical3D dataset enables the geometry transformer to learn robust 3D representations, markedly improving manipulation success rates.

\vspace{2pt}\noindent\textbf{Efficiency of Geometry Transformer.} We evaluate the inference latency of MASt3R and VGGT when used as the geometry transformer by running each policy 100 times on an RTX-4090 and reporting the average. MASt3R strikes a favorable balance between computational efficiency and task performance, achieving a latency of $56.2 ms$. In contrast, VGGT exhibits a significantly higher latency of $140.4$ ms, nearly $2.5\times$ slower. We observe that inference rates below 10 Hz introduce noticeable motion jitter, making VGGT model unsuitable for real-time deployment in surgical robotic systems.

\vspace{2pt}\noindent\textbf{Performance of MSFC.}
As shown in Fig.~\ref{fig:injector}, we also explore alternative designs for the spatial connectors.
\begin{itemize}
\item[--] \textit{Last-Layer Feature Connector (LFC).}
Following DINO~\cite{DINO2} and CLIP~\cite{CLIP}, where final-layer tokens encode the most semantic information, we inject tokens from the last geometry transformer layer by projecting them through a lightweight MLP to match the policy deocder’s feature dimension.
\item[--] \textit{Multi-Layer Separate Connector (MSC).}
Inspired by design of Pi-0~\cite{pi0}, we extract tokens from geometry transformer layers, project them to the policy deocder’s feature dimension, and perform separate cross-attention with action tokens in different transformer layers to incorporate multi-layer geometry.
\end{itemize}
As shown in Tab.~\ref{table:MGI}, LFC performs the worst as last-layer tokens provide limited spatial cues. MSC also underperforms, likely due to its complex attention design requiring large-scale data. In contrast, our proposed MSFC achieves the best results by efficiently fusing multi-level 3D latent embeddings into a compact representation, enabling robust policy learning from limited demonstrations.


\begin{table}[t]
\small
    \setlength{\abovecaptionskip}{0.3em}
    \centering
    \resizebox{\linewidth}{!}{
    \begin{tabular}{{l|cc|ccc}}
        \hline
        \hline
         \rowcolor[gray]{.92} & \multicolumn{2}{c|}{Peg Pickup} & \multicolumn{3}{c}{Knot Tying}\\
        \rowcolor[gray]{.92} Config  & test1 & test2 & Grasp & Loop & Whole Task\\
        \hline
        LFC & 1/10 & 0/10 & 0/10 & 0/10 & 0/10 \\
        MSC & 10/10 & 3/10 & 5/10 & 0/10 & 0/10\\
        \rowcolor{mycyan} MSFC(Ours) & 10/10 & 8/10 & 10/10 & 7/10 & 7/10\\
        \hline
        \hline
    \end{tabular}
    }
\caption{\textbf{Effectiveness of Multi-Level Spatial Feature Connector.} LFC: Last-Layer Feature Connector; MSC: Multi-Layer Separate Connector. MSFC: Multi-Level Spatial Feature Connector}
\vspace{-1em}
\label{table:MGI}
\end{table}
\section{Conclusion}
\label{conclusion}
In this paper, we propose Spatial Surgical Transformer (SST), an end-to-end visuomotor policy which is carefully designed for surgical robots to learn manipulation under the guidance of 3D spatial priors. Through comprehensive evaluations on three distinct real-world surgical tasks, we demonstrate that SST consistently achieves state-of-the-art performance comparable to existing methods, while remaining practical for clinical deployment. Our experimental results highlight that the key components of our framework, including finetuning geometry transformer on our Surgical3D dataset and the multi-level spatial feature connector, effectively empower surgical robots with spatial intelligence, thereby enhancing their reachability and generalization in surgical task. We hope this work will inspire further exploration toward autonomous surgery.

{
    \small
    \bibliographystyle{ieeenat_fullname}
    \bibliography{main}
}


\end{document}